\documentclass[10pt,twocolumn,letterpaper]{article}

\usepackage{wacv}
\usepackage{times}
\usepackage{epsfig}
\usepackage{graphicx}
\usepackage{amsmath}
\usepackage{amssymb}

\usepackage{pgfplots}
\usepackage{tikz}
\usepackage{caption}
\usepackage{subcaption}
\usepgfplotslibrary{colorbrewer}
\pgfplotsset{cycle list/Set1-9}
\tikzset{every picture/.style={line width=1pt}}
\usetikzlibrary{patterns}
\usepackage{booktabs}
\usepackage{graphicx}
\usepackage{subfiles}
\usepackage{amsmath,amssymb} 
\usepackage{algorithm}
\usepackage[noend]{algpseudocode}
\usepackage{bm}
\usepackage{color}
\usepackage[numbers]{natbib}
\usepackage{diagbox}
\newcommand{\C}{\mathcal{C}}
\newcommand{\Cbase}{\mathcal{C}_{\text{base}}}
\newcommand{\Cnovel}{\mathcal{C}_{\text{novel}}}
\newcommand{\SData}{\mathcal{S}}
\newcommand{\Strain}{\mathcal{S}_{\text{train}}}
\newcommand{\Stest}{\mathcal{S}_{\text{test}}}
\newcommand{\Strainnovel}{\mathcal{S}_{\text{train}}^{\text{novel}}}
\newcommand{\Sgennovel}{\mathcal{S}_{\text{gen}}^{\text{novel}}}

\wacvfinalcopy 

\def\wacvPaperID{571} 

\ifwacvfinal\fi
\begin{document}

\title{Self Paced Adversarial Training for Multimodal Few-shot Learning}

\author{Frederik Pahde\textsuperscript{1,2}, Oleksiy Ostapenko\textsuperscript{1,2},\\ Patrick J{\"a}hnichen\textsuperscript{2}, Tassilo Klein\textsuperscript{1}, Moin Nabi\textsuperscript{1} \\\\
\textsuperscript{1}SAP SE, Berlin, \textsuperscript{2}Humboldt-Universit{\"a}t zu Berlin\\
{\tt\small \{frederik.pahde, oleksiy.ostapenko, tassilo.klein, m.nabi\}@sap.com}\\
{\tt\small patricken.jaehnichen@hu-berlin.de}
}
\maketitle

\begin{abstract}
State-of-the-art deep learning algorithms yield remarkable results in many visual recognition tasks. However, they still fail to provide satisfactory results in scarce data regimes. To a certain extent this lack of data can be compensated by multimodal information. Missing information in one modality of a single data point (e.g. an image) can be made up for in another modality (e.g. a textual description). 
Therefore, we design a few-shot learning task that is multimodal during training (i.e. image and text) and single-modal during test time (i.e. image).
In this regard, we propose a self-paced class-discriminative generative adversarial network incorporating multimodality in the context of few-shot learning. The proposed approach builds upon the idea of cross-modal data generation in order to alleviate the data sparsity problem. We improve few-shot learning accuracies on the finegrained CUB and Oxford-102 datasets.



\end{abstract}

\section{Introduction}
The big data assumption is key for conventional deep learning applications but often also a limiting factor. 
However, in many applications it is too expensive (or even impossible) to acquire a sufficient number of training samples, resulting in inferior model accuracy. 
In contrast, humans are able to
quickly learn from only few instances. 
In consequence, research in the domain of few-shot learning, i.e. learning and generalizing from only few training samples, has gained more and more interest (e.g. \cite{ravi_optimization_2017,snell_prototypical_2017,vinyals_matching_2016}). 
However, research has focused on utilizing only one data modality (mostly images) so far. 
By including data from additional modalities (e.g. textual descriptions) we can overcome limitations in the low data regime, resulting in improved model performance. 
Our key assumption is that incorporating multimodal data (i.e. images and fine-grained descriptions thereof) forces the model to identify highly discriminative features \textit{across modalities} facilitating use in few-shot scenario.
Specifically, pursuing multimodality suggests that novel classes with low training data in one modality can benefit from previously learned features. 
\begin{figure*}[t!]
	\centering
  \includegraphics[width=.85\textwidth]{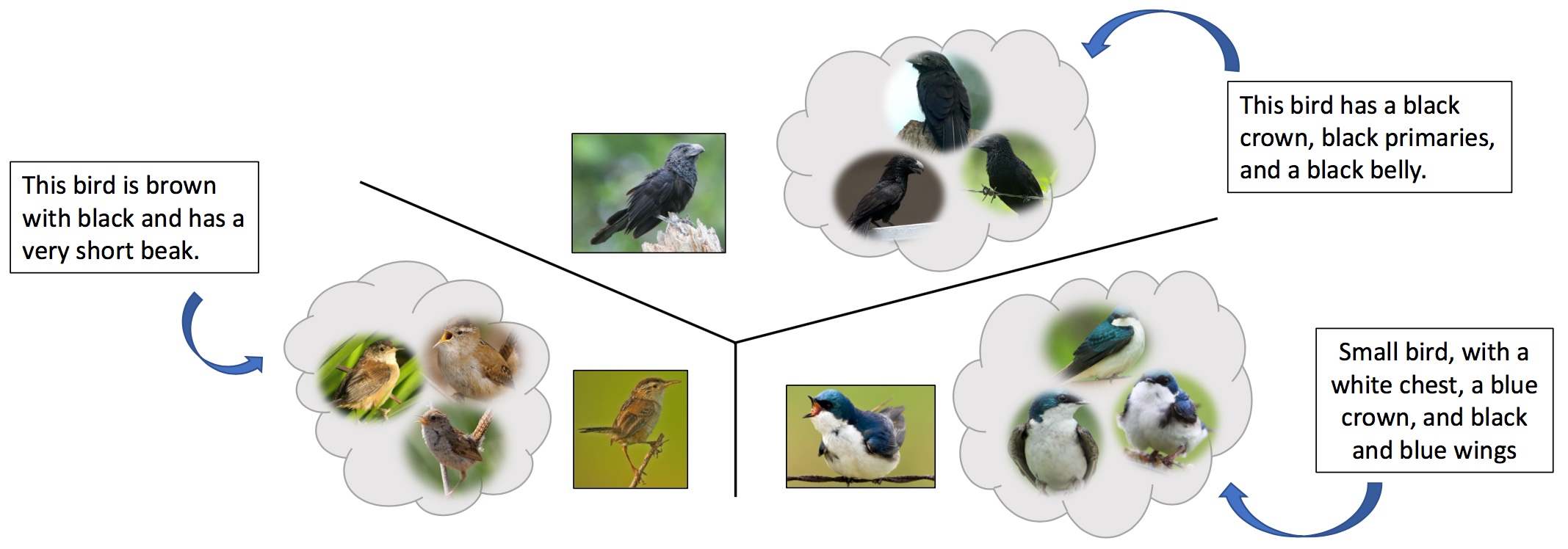}
	\caption{Learn classifier from dataset extended by hallucinated data conditioned on text}
	\label{fig:taskFigure}
\end{figure*}

In this regards, we propose to extend few-shot learning to incorporate multimodality in a meta-learning fashion. 
Specifically, we assume a scenario that is multimodal during training (i.e. images and texts) and single-modal during testing time (i.e. images). This comes with the associated task to utilize multimodal data in \emph{base} classes (with many samples) and to learn explicit visual classifiers for \emph{novel} classes (with few samples). This scenario for multimodal few-shot learning mimics situations that often arise in practice.
In such cases, cross-modal data hallucination is a viable solution. It facilitates few-shot learning by hallucinating images conditioned on fine-grained textual descriptions, and thus solving the scarce data problem. In this regard, Generative Adversarial Networks (GANs) have proven to be very effective for cross-modal sample generation. GANs~\cite{goodfellow_generative_2014,DBLP:journals/corr/MirzaO14} are deep networks mainly applied for unsupervised tasks and commonly used to generate data (e.g. images). The supervisory information in a GAN is indirectly provided within the frame of an adversarial game between two independent networks: a generator ($G$) and a discriminator ($D$).  During training, $G$ generates new data and $D$ tries to understand whether its input is real  (i.e. it is a training image) or it was generated by $G$. This competition between $G$ and $D$ is helpful in boosting the accuracy of both $G$ and $D$. At testing time, only $G$ is used to generate new data. Specifically, the {\em text-conditioned} GAN~\citep{reed16_gen, zhang_stackgan++:_2017,xu_attngan:_2017}, that we use in our approach, take as input a text and generate an image. Training the text-conditioned GAN allows for the generation of a potentially infinite number of samples given textual descriptions. However, the challenge is to pick adequate samples out of the pool of generated samples that allow for building a better classifier within the few-shot scenario. Such subset of images should not only be realistic but also class discriminative. To this end, we follow the \emph{self-paced strategy} and select a subset of images corresponding to ones in which the generator is most confident about their ``reality'' and the discriminator is the most confident about their ``class discriminativeness''. 
The main idea behind self-paced learning is that a subset of ``easy'' samples can be automatically selected in each iteration. Training is then performed using only this subset, which is progressively increased in the subsequent iterations when the model becomes more mature. Self-paced learning, applied in many other studies \cite{kumar2010self,sangineto2016self}, is related to curriculum learning \cite{bengio2009curriculum}, and is biologically inspired by the common human process of gradual learning, starting with the simplest concepts and increasing complexity. In this paper we adopt a self-paced learning approach to handle the uncertainty related to the quality of generated samples, thus ``easy'' is interpreted as ``high quality''. Specifically, a subset of ``high quality'' samples generated by the $G$ is automatically selected by the $D$ in each iteration, and training GAN is then performed using only this subset. Intuitively, we select a subset of the generated samples that the classifier trained on the real data is most confident about, as illustrated in Fig. \ref{fig:taskFigure}.  

The main contributions of this work are three-fold: 
\textbf{First}, we extended the few shot learning setting proposed by Hariharan and Girhsick~\cite{hariharan_low-shot_2017} to work with multimodal data during training time.
\textbf{Second}, we propose a class-discriminative text-conditional GAN that facilitates few-shot learning by hallucinating additional training images. 
\textbf{Third}, we leverage a self-paced learning strategy facilitating reliable cross-modal hallucination.  
Our approach features robustness and outperforms the baseline in the challenging low-shot scenario.


\section{Related Work}
In this section we briefly review previous work considering: (1) few-shot learning, (2) multimodal learning and (3) self-paced learning.

\subsection{Few-Shot Learning}
For learning deep networks using limited amounts of data, different approaches have been developed. Following Taigman et al.~\cite{taigman2014deepface}, Koch et al.~\cite{koch_siamese_2015} interpreted this task as a verification problem, i.e. given two samples, it has to be verified, whether both samples belong to the same class. Therefore, they employed siamese neural networks \cite{bromley1994signature} to compute the distance between the two samples and perform nearest neighbor classification in the learned embedding space. Some recent works approach few-shot learning by striving to avoid overfitting by modifications to the loss function or the regularization term. Yoo et al.~\cite{yoo_efficient_2017} proposed a clustering of neurons on each layer of the network and calculated a single gradient for all members of a cluster during the training to prevent overfitting. The optimal number of clusters per layer is determined by a reinforcement learning algorithm. A more intuitive strategy is to approach few-shot learning on data-level, meaning that the performance of the model can be improved by collecting additional related data. Douze et al.~\cite{douze_low-shot_2017} proposed a semi-supervised approach in which a large unlabeled dataset containing similar images was included in addition to the original training set. This large collection of images was exploited to support label propagation in the few-shot learning scenario. Hariharan et al.~\cite{hariharan_low-shot_2017} combined both strategies (data-level and algorithm-level) by defining the squared gradient magnitude loss, that forces models to generalize well from only a few samples, on the one hand and generating new images by hallucinating features on the other hand. For the latter, they trained a model to find common transformations between existing images that can be applied to new images to generate new training data (see also \cite{wang_low-shot_2018}). Other recent approaches to few-shot learning have leveraged meta-learning strategies. Ravi et al.~\cite{ravi_optimization_2017} trained a long short-term memory (LSTM) network as meta-learner that learns the exact optimization algorithm to train a learner neural network that performs the classification in a few-shot learning setting. This method was proposed due to the observation that the update function of standard optimization algorithms like SGD is similar to the update of the cell state of a LSTM. Bertinetto et al.~\cite{bertinetto_learning_2016} trained a meta-learner feed-forward neural network that predicts the parameters of another, discriminative feed-forward neural network in a few-shot learning scenario. Another tool that has been applied successfully to few-shot learning recently is attention. Vinyals et al.~\cite{vinyals_matching_2016} introduced matching networks for one-shot learning tasks. This network is able to apply an attention mechanism over embeddings of labeled samples in order to classify unlabeled samples. One further outcome of this work is that it is helpful to mimic the one-shot learning setting already during training by defining mini-batches, called episodes with subsampled classes. Snell et al.~\cite{snell_prototypical_2017} generalize this approach by proposing prototypical networks. Prototypical networks search for a non-linear embedding space (the prototype) in which classes can be represented as the mean of all corresponding samples. Classification is then performed by finding the closest prototype in the embedding space. In the one-shot scenario, prototypical networks and matching networks are equivalent. 

\subsection{Multimodal Learning}
\cite{kiros_unifying_2014} propose to align visual and semantic information in a joint embedding space using a encoder-decoder pipeline.
Building on this, \cite{faghri_vse++:_2017} improve upon this mixed representation by incorporating a triplet ranking loss. 

\cite{karpathy_deep_2015} generate textual image descriptions. Their model infers latent alignments between regions of images and segments of sentences of their respective descriptions. 
\cite{reed_learning_2016} focus on fine-grained visual descriptions. 
They present an end-to-end trainable deep structured joint embedding trained on two datasets containing fine-grained visual descriptions. 

In addition to multimodal embeddings, another related field using data from different modalities is text-to-image generation. 
\cite{reed16_gen} study image synthesis based on textual information. \cite{zhang_stackgan++:_2017} greatly improve the quality of generated images to a photo-realistic high-resolution level by stacking multiple GANs (StackGANs). 
Extensions of StackGANs include an end-to-end trainable version \cite{zhang_stackgan++:_2017} and considering an attention mechanism over the textual input \citep{xu_attngan:_2017}. Sharma et al. \cite{sharma2018chatpainter} extended the conditioning by involving dialogue data and further improved the image quality.
Beside the usage of GANs for conditioned image generation, other work employed Variational Autoencoders \cite{kingma2013auto} to generate images \cite{mishra2017generative}. However, they conditioned on attribute vectors instead of text.

\subsection{Learning from Simple to Complex}
Recently, many studies have shown the benefits of organizing the training examples 
from simple to complex for model training. Bengio et al. \cite{bengio2009curriculum} first proposed a general learning strategy: curriculum learning. They show that suitably sorting the training samples, from the easiest to the most difficult, and iteratively training a classifier starting with a subset of easy samples 
can be useful to find better local minima. In \cite{chen2015webly}, easy and difficult images 
are provided for training a CNN in order to learn generic CNN features using webly annotated data. Note that in this and in all the other curriculum-learning-based approaches, the order of the samples is provided by an external supervisory signal, taking into account human domain-specific expertise. 

Curriculum learning was extended to self-paced learning by Kumar et al. \cite{kumar2010self}. They proposed the self-paced learning framework, automatically expanding the training pool in an easy-to-hard manner by converting the curriculum mechanism into a concise regularization term. Curriculum learning uses human design to organize the examples, and self-paced learning can automatically choose training examples according to the loss. Supancic et al. \cite{supancic2013self} adopt a similar framework in a tracking scenario and train a detector using a subset of video frames, showing that this selection is important to avoid drifting.
In \cite{zhang2017bridging} saliency is used to progressively select samples in weakly supervised object detection.

Although some of these self-paced methods use pre-trained CNN-based features to represent samples (e.g., \cite{liang2015towards}), or uses a CNN as the classifier directly(e.g., \cite{sangineto2016self}), none of them formulates the self-paced strategy in a GAN training protocol as we do in this paper.

\begin{figure*}[t]
	\centering
  \includegraphics[width=0.95\textwidth]{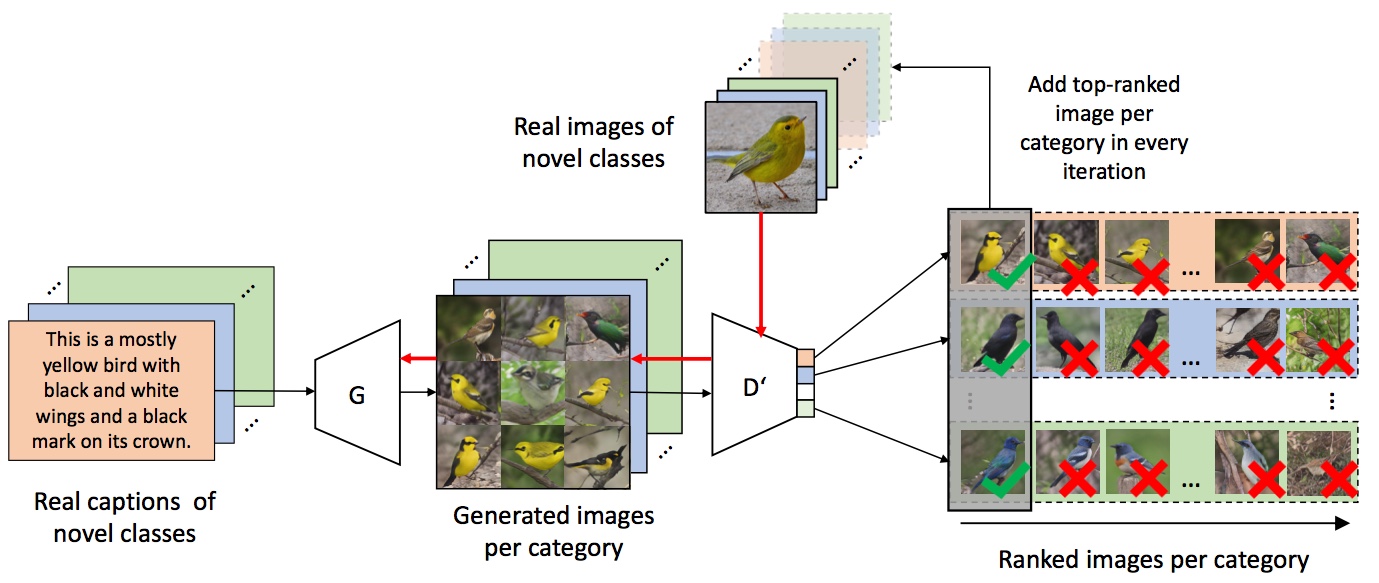}
	\caption{Single iteration of self-paced finetuning on novel classes: $G$ generates images, $D'$ ranks the generated images based on their class-discriminatory power. Then the ``best'' images are added to the real samples and used to update $D'$ and $G$. This process is repeated multiple times.}
	\label{fig:method}
\end{figure*}
\section{Background on GANs}
\label{sec:background}

\textbf{GAN:} 
GANs \cite{goodfellow_generative_2014} consist of two adversarial agents that compete with each other during a Minimax game until reaching a Nash equilibrium at convergence of the training phase. 
Specifically, they decompose into two main components. 

First, a \emph{generator} $G$ aims to produce \emph{fake} image data $i_{fake}$ resembling \emph{real} data. That is $i_{fake} = G(z)$, given a random noise vector $z$, which induces a generator distribution $p_G$. Second, a \emph{discriminator} $D$ that defines a mapping $D: \mathcal{I}\to \left[0,1\right]$ that given an image $i\in I$, outputs a distribution w.r.t. state $s$ (i.e. \emph{fake} or \emph{real}), $p(s\mid i)=D(i)$. To this end, the GAN parameters are obtained by optimization of the loss function $\mathcal{L}\left(.\right)$,
\begin{eqnarray}
\label{stdGANloss}
\mathcal{L}_{GAN}\left(G,D\right)=\mathbb{E}_{I}\left[\log D\left(I\right)\right]\\ \nonumber +\mathbb{E}_{z}\left[\log D\left(I,G\left(z\right)\right)\right].
\end{eqnarray}
Specifically, training the GAN entails alternating optimization of $D$ and $G$ using stochastic gradient-based descent.
\cite{goodfellow_generative_2014} prove that a global optimium is reached when $p_g = p_{data}$, and under mild conditions  eventually $p_g$ converges to $p_{data}$.

\textbf{Conditional GAN:}
Conditional GANs (cGANs)~\cite{DBLP:journals/corr/MirzaO14} constitute an extension of the standard GAN, where the input of the generator as well as discriminator are further tied to a conditional variable containing extra information. That condition variable be an kind of auxiliary information such as a class label \cite{DBLP:journals/corr/MirzaO14}, an attribute vector \cite{yan2016attribute2image} or a semantic description~\cite{reed16_gen}. To this end, the standard loss Eq.~\ref{stdGANloss} changes slightly to
\begin{eqnarray}
\label{cGANloss}
\mathcal{L}_{GAN}\left(G,D\right)=\mathbb{E}_{I}\left[\log D\left(I\mid y \right)\right]\\ \nonumber +\mathbb{E}_{z}\left[\log D\left(I,G\left(z\mid y\right)\right)\right],
\end{eqnarray}
where the conditional variable is denoted as $y$. Technically, conditioning is commonly achieved by concatenation of variables, although there exist numerous possibilites - see~\cite{KwakZ16a} for some examples.
\section{Method}

\subsection{Preliminaries}
Before developing our method, we introduce the necessary notation.
Let $\mathcal{I}$ denote the image space, $\mathcal{T}$ the text space and $\C=\lbrace 1,...,R\rbrace$ be the discrete label space. Further, let $x_i \in \mathcal{I}$ be the $i$-th input data point, $t_i \in \mathcal{T}$ its corresponding textual description and $y_i \in \C$ its label.
In the few-shot setting, we consider two disjunct subsets of the label space: $\Cbase$, labels for which we have access to sufficient  data samples; and novel classes $\Cnovel$ which are underrepresented in the data. Note that both subsets exhaust the label space $\C$, i.e. $\C = \Cbase \cup \Cnovel$. We further assume that in general $|\Cnovel| < |\Cbase|$.

We organize the data set $\SData$ as follows.
Training data $\Strain$ consists of tuples $\{(x_i, t_i, y_i)\}_{i=1}^{n}$ taken from the whole data set and test data $\Stest = \{(x_i, y_i) : y_i \in \Cnovel\}_{i=1}^m$ that belongs to novel classes and $\SData = \Strain \cup \Stest$, $\Strain \cap \Stest = \emptyset$.
Naturally, we can also consider $\Strainnovel = \{(x_i, t_i, y_i) : (x_i, t_i, y_i) \in \Strain, y_i \in \Cnovel\}_{i=1}^k \subset \Strain$,
where in accordance with a few-shot scenario $k = \left|\Strainnovel\right|\ll\left|\Strain\right| = n$. 
Additionally, in a few-shot learning scenario, the number of samples per category of $\Cbase$ may be limited to $g$, denoted by $\Strainnovel(g)$.

\subsection{Text-conditioned Data Generation}    
The core idea of our method is to improve accuracy in few-shot learning scenarios by using an augmented dataset with additional hallucinated samples conditioned on textual descriptions. 
For that purpose we employ a text-conditioned GAN (tcGAN) \citep[e.g.][]{reed16_gen, zhang_stackgan++:_2017,xu_attngan:_2017} which can be interpreted as a variant of cGAN - see Sec.~\ref{sec:background} for details.

The purpose of a tcGAN is to learn the mapping $G:\mathcal{T}\to \mathcal{I}$. In this regard, $G$'s objective is to generate samples $i\in\mathcal{I}$ conditioned on textual descriptions $t\in\mathcal{T}$ that cannot be distinguished from ``real'' images. In contrast, the adversarially trained discriminator $D$ aims to detect generated ``fake'' samples. To do so, $D$ generates two probability distributions: $D_s(i) = p(s|i)$, a distribution over the state of the image (``real'' or ``fake''), and $D_t(i) = p(t|i)$, a distribution over textual representations\footnote{Note that we are implicitly using text embeddings as a textual representation.} for a given image $i$.

Slightly abusing notation, let $T = \{t_1, \ldots, t_n\}$, $I = \{i_1,\ldots,i_n\}$ be observed texts, images and classes respectively.
The objective of a tcGAN can then be expressed as
\begin{eqnarray}
\label{tcGANloss}
\mathcal{L}_{tcGAN}\left(G,D\right)=\mathbb{E}_{I,T}\left[\log D_T\left(I\right)\right]\\\nonumber +\mathbb{E}_{T,z}\left[\log D_S\left(G\left(T,z\right)\right)\right],
\end{eqnarray}
where $z$ denotes a random noise vector.
\newline
We use the StackGAN architecture proposed by \cite{zhang_stackgan++:_2017} as our tcGAN. Here, the idea is to use multiple GANs with different levels of granularity. In a StackGAN with $l$ stacked GANs, we consider generators $G_1, \ldots, G_l$ and discriminators $D_1, \ldots, D_l$. 
Now, $G_1$ is conditioned on a text embedding $\varphi_t$ for text $t$ and generates a low-resolution image $i_1$. 
Both the generated image $i_1$ and $\varphi_t$ act as input to $D_1$ which in turn predicts whether the image is real or fake given the textual description. 
On a next stage, $G_2$ takes the generated image provided from $G_1$ in conjuction with the textual embedding as input in order to generate a more detailed image of higher resolution.
Having this pipeline, the image quality is increased at every stage of the StackGAN, resulting in a high-resolution image at its final stage. See \cite{zhang_stackgan++:_2017} for further details.\\
StackGANs allow for text-conditioned image synthesis optimized for realistic appearance. 
However, they lack the ability to take into account that textual representations and images might be labeled with class information. This calls out for an extension to utilize class labels in a augmented-with-hallucinated-data few-shot scenario and is presented next.

\subsection{Auxiliary Classifier GAN}
\label{subsec:auxclassifier}
Conventional tcGANs cannot consider class labels 
and are therefore not adequate in the few-shot scenario.
Hence, we propose to employ the auxiliary classifier GAN architecture \cite{odena2016conditional}. 
Specifically, this entails augmentation of the default tcGAN objective as in Eq.~\eqref{tcGANloss} with a classification loss $\mathcal{L}_{class}$, which is defined as
\begin{equation}
\mathcal{L}_{class}\left(D\right)=\mathbb{E}_{C,I}\left[\log p\left(C\mid I\right)\right].
\end{equation}
Further, let
\begin{equation*} \mathcal{L}_{class}\left(G\right)\triangleq\mathcal{L}_{class}\left(D\right).
\end{equation*}
Now, augmenting the objective leads to the two loss terms,
\begin{equation}\mathcal{L}\left(D\right)=\mathcal{L}_{tcGAN}\left(G,D\right)+\mathcal{L}_{class}\left(D\right)
\end{equation}
\begin{equation}\mathcal{L}\left(G\right)=\mathcal{L}_{tcGAN}\left(G\right)-\mathcal{L}_{class}\left(G\right),
\end{equation}
which are optimized in an alternating fashion.\\
Conceptually, optimization of the augmented loss implies that $D$ solves the additional task of predicting the class label of given images in addition to discrimination between ``real'' or ``fake''. 
Adding the auxiliary output layer to the StackGAN architecture therefore suggests a two-fold advantage in the context of few-shot learning. 
First, backpropagating the classification loss to $G$ favors the generation of samples which are both class-specific and realistic. 
This will prove to be key for performing classification using a dataset extended with generated (i.e. hallucinated) samples. 
From now on, we denote images generated by $G$ for $\Cnovel$ as $\Sgennovel$. 
Second, the new output layer of $D$ can be used to perform classification. 
As a consequence, $D$ can be used as a classifier both for novel classes and base classes for which meaningful latent representations are readily available. 

\begin{algorithm}[ht]
	\caption{Self-paced adversarial training, \textsc{RANK}() is a function that ranks generated images based on their score of $D'$ and \textsc{TOP}() returns the highest ranked images}\label{spl}
\begin{algorithmic}[1]
	\State \textbf{Input:} Pre-trained networks $G$, $D'$ and $K$
	\State \textbf{Output:} Finetuned classifier $D'$
	\For{$i = 1,\ldots, n$}
	\State $\Sgennovel = \emptyset$
	\For{$c \in \Cnovel$}
    	\State $\text{candidates}=\emptyset$
        \For{$\text{caption} \in t_c$}
        	\State $\text{candidates} = \text{candidates}\cup G(\text{caption})$
        \EndFor
		
		\State $\text{candidates}_{\text{ranked}}$ $=$ \textsc{rank}$(\text{candidates}, D')$
		\State $\text{sample} = \textsc{top}(\text{candidates}_{\text{ranked}}, K)$
		\State $\Sgennovel = \Sgennovel \cup \text{sample}$
	\EndFor
	\State $\mathcal{S}_{\text{all}}^{\text{novel}} = \Strainnovel \cup \Sgennovel$
	\State update $D'$, $G$ with $\mathcal{S}_{\text{all}}^{\text{novel}}$
	\EndFor
\end{algorithmic}
\end{algorithm}
\subsection{Self-paced Finetuning on Novel Classes}
The representation learning phase, which consists of training the StackGAN with an auxiliary classifier, yields the discriminator $D$. 
As described, the discriminator is able to distinguish real from fake samples as well as to perform classification w.r.t. the base classes. 
However, to classify w.r.t. novel classes as well, $D$ has to be adapted. 
Specifically, the class-aware layer with $|\Cbase|$ output neurons is replaced and reduced to $|\Cnovel|$ output neurons, which are randomly initialized. 
We refer to this modified discriminator as $D'$. 
Now, using the notion of hallucinating additional samples, the network can be finetuned with generated images as well as training data for novel categories, i.e. with the data given by $S_{train}^{novel} \cup S_{gen}^{novel}$. 
It should be noted that samples contained in $\Sgennovel$ can be very noisy which can be attributed to the fact that $G$ does not always output high-quality images. 
In order to alleviate that problem, we propose a self-paced learning strategy ensuring that only the best generated samples within $\Sgennovel$ are used. 
In particular, we employ the softmax activation for class-specific confidence scoring. 
To this end, optimization is performed on the loss defined as, 
\begin{eqnarray}
\max_{G} \min_{D, \boldsymbol{\alpha}} \mathcal{L}\left(D,G \mid I_{novel}, T_{novel} \right)\\\nonumber  + \sum_{T \in {T}_{\text{novel}}} \alpha_{T} \mathbb{E}_{I_T \sim G(T) } [ \log D( I_T ) ] \nonumber 
\label{eq:2} 
\end{eqnarray}
\begin{eqnarray}
\text{subject to:}  & 0 \leq \alpha_T \leq 1, \quad \sum_{T \in {T}_{\text{novel}}} \alpha_I \leq K, \nonumber 
\end{eqnarray}

where $\mathcal{L}\left(D,G \mid  I_{novel}, T_{novel} \right) $ is the auxiliary GAN loss on the joint set of generated and training data $\Strainnovel \cup \Sgennovel$,
$G(T)$ is the tcGAN generator, and $\alpha_T$ is a soft selector for the images $I_T$ generated from textual descriptions $T$, where the descriptions come from a set of captions for novel categories $T_{novel}$ (i.e. $T \in {T}_{\text{novel}}$). 
Since each $\alpha_T$ is a selector for the generated data, $K$ specifies 
the maximum number of generated
of samples to be included in $\Sgennovel$ for the next finetuning step. Our pseudo-code is given in algorithm~\ref{spl}.

\subsubsection{Initialization for SPL}
To obtain a meaningful ranking in the self-paced learning phase, $D'$ has to be initialized on novel classes. 
Again taking into account the setting of few-shot learning, we restrict the number of samples per class available to $n$ for doing this, i.e. use $\Strainnovel(n)$ instead of $\Strainnovel$.
Since only real images are used for initialization, the quality of this data is very high compared to the noise-prone generated samples. 
Due to the limited amount of samples, the initialized $D'$ will be weak on the classification task, but sufficiently powerful for performing an initial ranking of the generated images.

\subsubsection{Self-paced Adversarial Training}
The ability to rank generated images with the pre-trained $D'$ allows for data selection and guided optimization. 
In our approach we specifically follow a self-paced learning strategy. 
This entails iteratively choosing generated images that have highest probability in $D'$ for $\Cnovel$, yielding a curated set of high-quality generated samples $\Sgennovel$. 
Finally, we aggregate original samples and generated images $\Strainnovel \cup \Sgennovel$ for training, during which we alternately update $D'$ and $G$. 
Doing so yields both a more accurate ranking as well as higher class prediction accuracy as the number of samples increases.
Ultimately, the approach summarized in algorithm~\ref{spl} learns a reliable classifier that performs well in few-shot learning scenarios.

\section{Experiments}
\begin{table}
    \setlength{\tabcolsep}{5pt}
	\centering{
		\begin{tabular}{lccccc} \toprule
			& & & n	& 	&\\
			Model  & 1 & 2 & 5 & 10 & 20\\ \midrule
             Finetuning  	& 40.79	& 39.47 & 56.07	& 66.25	& 73.08	\\
            Initialization 	& 49.16	& 58.02 & 69.6 & 77.89 & 82.98 \\
            SPL-$D'$  	& 55.16 & 58.16 & 71.41 & \textbf{78.52} & 84.03 \\
            SPL-$D'$G  	& \textbf{57.67} & \textbf{59.83} & \textbf{73.01} & 78.1 & \textbf{84.24} 
			 \\\bottomrule
		\end{tabular}
	}
	\captionsetup{position=above}
	\captionof{table}{Top-5 accuracy in percent for our model on the CUB dataset in different settings (best results are in bold)}
	\label{tab:results}
\end{table}
\subsection{Datasets}
We test the applicability of our method on two fine-grained classification datasets, namely CUB \cite{WahCUB_200_2011} with bird data and Oxford-102~\cite{nilsback2008automated} containing flower data.
Specifically, the CUB bird dataset contains 11,788 images of 200 different bird species, with $\mathcal{I} \subset \mathbb{R}^{256\times256}$. 
The data is split equally into training and test data. 
As a consequence, samples are roughly equally distributed, with training and test each containing $\approx 30$ images per category. 
Additionally, 10 short textual descriptions per image are provided by \cite{reed_learning_2016}. 
Similar to \cite{zhang_stackgan++:_2017}, we use the text-encoder pre-trained by \cite{reed_learning_2016}, yielding a text embedding space $\mathcal T \subset \mathbb{R}^{1024}$. 
Following \cite{zhang_stackgan++:_2017}, we split the data such that $\left|C_{base}\right|=150$ and $\left|C_{novel}\right|=50$. 
To simulate few-shot learning, $n\in\{1,2,5,10,20\}$ images of $C_{novel}$ are used for training, as proposed by \cite{hariharan_low-shot_2017}.
In contrast, the Oxford-102 dataset contains images of 102 different categories of flowers. Similar to the CUB-200 dataset, Reed et al.~\cite{reed_learning_2016} provide 10 textual descriptions per image. As for the CUB dataset, we use the text-encoder pre-trained by \cite{reed_learning_2016}, yielding a text embedding space $\mathcal T \subset \mathbb{R}^{1024}$. Following Zhang et al.~\cite{zhang_stackgan++:_2017}, we split the data such that $\left|C_{base}\right|=82$ and $\left|C_{novel}\right|=20$. 
To simulate few-shot learning, $n\in\{1,2,5\}$ images of $C_{novel}$ are used for training.

\begin{table}[t!]
    \setlength{\tabcolsep}{5pt}
	\centering{
		\begin{tabular}{lccc} \toprule
			&  & n	& 	\\
			Model  & 1 & 2 & 5\\ \midrule
             Finetuning  	& 70.42 &	82.53	& 	81.14 \\
Initialization  	& 75.26	& 89.45	& 89.97	\\
             SPL-D'G 	& \textbf{78.37}	& \textbf{91.18}	& \textbf{92.21}
            
			 \\\bottomrule
		\end{tabular}
	}
	\captionsetup{position=above}
	\captionof{table}{Top-5 accuracy in percent for our model on Oxford-102 dataset in different settings (best results are in bold)}
	\label{tab:results_oxford}
\end{table}
\begin{figure*}[ht]
	\centering
	\captionsetup{justification=centering, margin=-1cm}
    \begin{tikzpicture}[scale = 1]%
			\begin{axis}[
            legend columns=2,
            width=0.7\textwidth,
       		height=0.4\textwidth,
			legend pos=south east,
            major grid style={line width=.1pt,draw=gray!40},
            grid=both,
			xtick={0,3,6,...,30},
            ytick={0.45, 0.475 ,0.5,...,0.6},
            ymin=0.42,
            xlabel={Iteration in self-paced learning phase},
            ylabel= Accuracy,
            axis y discontinuity=crunch,
            xmin=0,xmax=30,
            ymax=0.6]
            \addplot+[mark=none] table [x=iteration, y=top5, col sep=comma] {data/evolution_DG.csv};
            \addplot+[mark=none] table [x=iteration, y=top5, col sep=comma] {data/evolution_D.csv};
            \addplot+[domain=0:30, mark=none, dashed]{0.4916};
            \legend{D' and G, D', Initialization}
            \end{axis}
		\end{tikzpicture}%
	
	\caption{ Top-5 Accuracy in every iteration for different strategies in the 1-shot learning scenario}
    \label{fig:evolution}
\end{figure*}
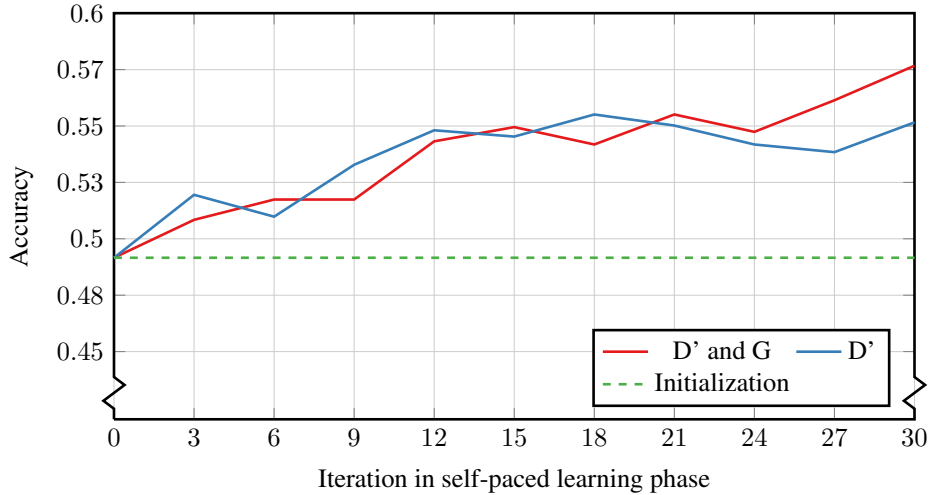

\subsection{Algorithmic Details}
During representation learning, we train a StackGAN for $900$ epochs. Similar to \cite{zhang_stackgan++:_2017}, we use Adam \cite{kingma2014adam} for optimization.
We set the learning rate $\tau$ to $2 \times 10^{-4}$ and the batch size to 24 for both $G$ and $D$. 
In the initialization phase for self-paced learning, we construct $D'$ by replacing the last layer of $D$ by a linear softmax layer. 
The resulting network is then optimized using the cross-entropy loss function and a SGD optimizer with learning rate $\tau=10^{-3}$ and momentum $0.5$. 
Batch size is set to 32 and training proceeds for 20 epochs. 
Self-paced learning of $D'$ continues to use the same settings (i.e. SGD with $\tau=10^{-3}$ and momentum $0.5$, minimizing a cross-entropy loss). 
Additionally, Adam's learning rate for $G$ is reduced to $2 \times 10^{-5}$. 
In every iteration we choose exactly one generated image per category and perform training for 10 epochs.
\begin{table}
    \setlength{\tabcolsep}{5pt}
	\centering{
		\begin{tabular}{lcccc} \toprule  
                  & Top 1  & Top 3  & Top 5 \\ \midrule
SGM \cite{hariharan_low-shot_2017}               & 19.08  & 40.55  & 48.89  \\
SGM+Hallucination \cite{hariharan_low-shot_2017} & 20.27  & \textbf{41.06}  & 50.59  \\
Our proposed method         & \textbf{24.90}  &  37.59  & \textbf{57.67}
			 \\\bottomrule
		\end{tabular}
	}
	\captionsetup{position=above}
	\captionof{table}{Top-1, top-2 and top-5 accuracy of single modality models for 1-shot Learning task proposed by \cite{hariharan_low-shot_2017}(SGM-loss and SGM-loss with Hallucination) compared to our model (best results are in bold) }
	\label{tab:comparison}
\end{table}
\subsection{Models}
In order to asses the performance of individual components, we perform an ablation study. 
A simple approach for transfer learning is to make use of a pre-trained representation and then finetune that network on novel data. 
We apply this strategy on a first baseline (\textbf{Finetuning}), for which we pre-train a classifier $T$ that has exactly the same architecture as $D$ on the base classes, followed by finetuning with the few instances of novel classes on $\Strainnovel$. 
This meta-learning strategy learns meaningful representations on the base classes $\Cbase$ that can be used for novel classes $\Cnovel$. 
A second baseline (\textbf{Initialization}) constitutes our first contribution. 
We modify the discriminator $D$ of the StackGAN, which we obtain from the representation learning phase, to obtain $D'$ by exchanging the discriminator's last layer. 
Finetuning is then performed on the real samples from novel classes $\Strainnovel$. 
Note that the \textit{initialization} baseline uses $D$ which is pre-trained using the adversarial principle during the StackGAN training, in contrast to the \textit{finetuning} baseline that uses $T$ as pre-trained by a conventional classifier.  
Afterwards, we iteratively add high-quality generated samples for  novel categories $\Sgennovel$ as described. 
In a first self-paced experiment (\textbf{SPL-D'}) we update $D'$ using selected generated samples in every iteration. 
In a second experiment, additionally to updating $D'$, we update $G$ in every iteration in order to be fully self-paced (\textbf{SPL-D'G}). Following previous approaches (e.g. \citep{hariharan_low-shot_2017}), we evaluate our approach by reporting the top-5 accuracy. 

\begin{figure*}[ht]
	\centering
  \includegraphics[width=0.95\textwidth]{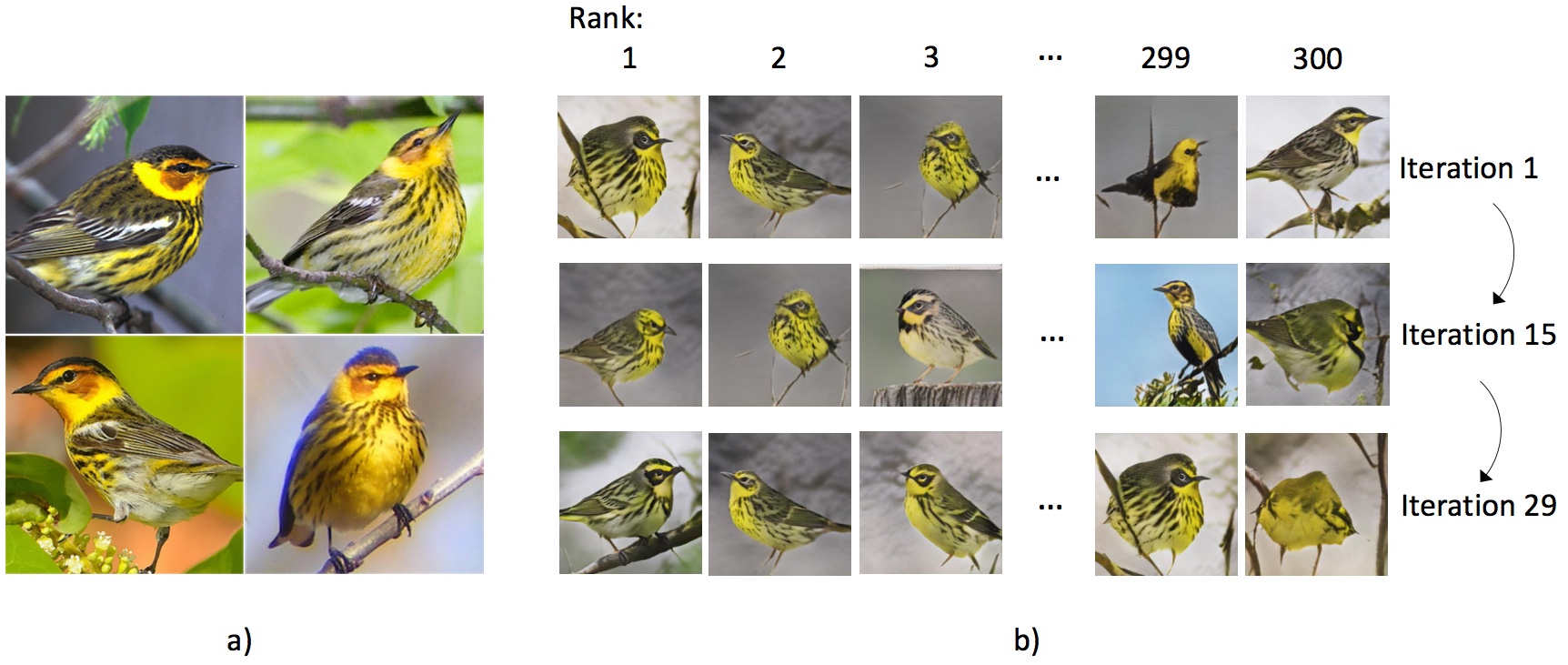}
	\caption{a) real images of birds and b) ranked birds after different iterations}
	\label{fig:lastFigure}
\end{figure*}
\subsection{Analysis of Self-paced Finetuning}
We run several additional experiments to further analyze the behavior of our method. For the following experiments we use the CUB bird dataset. 
\subsection{Results of Ablation Study}
\subsubsection{CUB-200 Dataset}
We report top-1-, top-3- and top-5-accuracy for our method in different settings. 
The results in top-5-accuracy are shown in Tab.~\ref{tab:results}. 
We observe that the initialization phase already provides a large margin compared to the finetuning phase. 
Both are finetuned exclusively on real images with the difference that our initialized $D'$ is pre-trained in an adversarial fashion on text and image data during the representation learning phase. 
In contrast, the finetuning baseline is pre-trained only on image data without adversarial training. 
Our results indicate that including text information in representation learning already provides accuracy gains. 
Further, we observe that the self-paced finetuning phase improves classification accuracy. 
In particular, the margin in a 1-shot scenario is large. 
For most scenarios, by updating both $G$ and $D'$ we achieve higher accuracy than by only updating $D'$. 
This observation indicates that generated images add upon class-discriminatory power through our self-paced finetuning strategy if $G$ is involved in training. 
Our full self-paced sample selection procedure, including updating $D'$ and $G$ provides an accuracy boost of $17$, $20$ and $17$ percent in the challenging $1$-,$2$- and $5$-shot scenarios, respectively. 
Also for $10$- and $20$-shot learning we outperform the baseline by more than $10$ percent.
We observe the same trends for top-1- and top-3 accuracy.

\subsubsection{Oxford-102 Dataset}
 Similar to the experiments for CUB, we compare our method (1) to a \textit{Finetuning} baseline, i.e. pretrain classifier on base classes and finetune it on novel classes and (2) to an \textit{Initialization} model, i.e. finetune $D^*$ obtained from representation learning on novel classes. For our method, we update both $D'$ and $G$ (SPL-D'G). Classification results in top-5-accuracy are given in Tab. \ref{tab:results_oxford}. We can observe similar trends compared to the CUB dataset. In all few-shot scenarios, we outperform the finetuning baseline with a large margin. Specifically, our method yields a performance boost by ca. 8\%, 10.5\% and 11\% in the challenging 1-, 2- and 5-shot scenarios compared to the finetuning baseline. The initialization for the self-paced learning phase provides performance gains as well. However, the full model yields the best results in every few-shot scenario.

\subsection{Comparison to Similar Work}
We compare our method to the single modality methods recently proposed by Hariharan and Girshick \cite{hariharan_low-shot_2017}. For the comparison in Tab.~\ref{tab:comparison} we use the CUB bird dataset. It can be observed that our approach outperforms the methods proposed by \cite{hariharan_low-shot_2017}, justifying the usage of multimodal data for training in few-shot scenarios.

\subsubsection{Evolution of classifiers during SPL}
In order to check the behavior of the the self-paced learning phase, we study the classification accuracy across all iterations. In Fig. \ref{fig:evolution} we report the top-5-accuracy for 30 iterations in the 1-shot learning scenario. The dashed line shows the accuracy after the initialization phase and can be interpreted as lower bound for the self-paced learning phase. The blue line shows the accuracy for our method updating only $D'$, while the red line shows the accuracy for updating $D'$ and $G$. For the first iterations both models behave very similarly, but after a certain warm-up phase the model entailing updates of $G$ performs better. That indicates that updating $G$ leads to the synthesis of images of higher quality, which in turn further increases the classification accuracy.

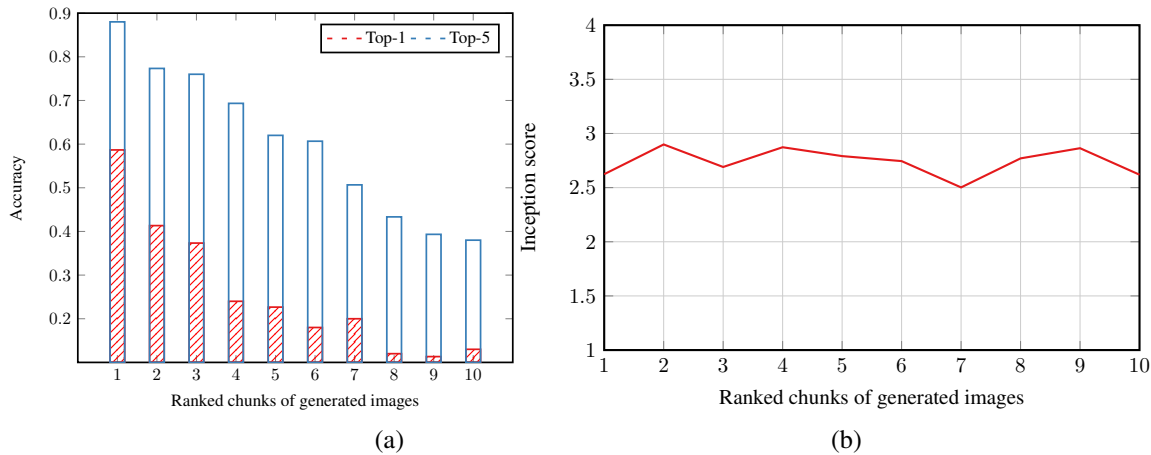
\begin{figure*}[ht]
	\centering
	\captionsetup{justification=centering, margin=-1cm}
	\begin{tikzpicture}[scale = 0.65]%
	\begin{axis}[
	legend pos=north east,
	legend columns=2,
	width=0.6\textwidth,
	height=0.5\textwidth,
	major grid style={line width=.1pt,draw=gray!40},
	grid=none,
	xtick={1,2,...,10},
	ytick={0.2,0.3,0.4,...,1.0},
	ymin=0.1,
	xlabel={Ranked chunks of generated images},
	ylabel= Accuracy,
    bar width=0.3cm,
	ymax=0.9,xmax=11]
    \addplot+[ybar, mark=none, , pattern=north east lines,pattern color = red] table [x=step, y=top1, col sep=comma] {data/rankingAccuracy.csv};
	\addplot+[ybar, mark=none] table [x=step, y=top5, col sep=comma] {data/rankingAccuracy.csv};
	
	\legend{Top-1, Top-5}
	\end{axis}
	\end{tikzpicture}%
    \begin{tikzpicture}[scale = 0.8]%
			\begin{axis}[
            legend columns=2,
            width=0.6\textwidth,
       		height=0.4\textwidth,
			legend pos=south east,
            major grid style={line width=.1pt,draw=gray!40},
            grid=both,
			xtick={0,1,...,10},
            ytick={1.0, 1.5 ,2.0,...,4.0},
            ymin=1,
            xlabel={Ranked chunks of generated images},
            ylabel= Inception score,
            xmin=1,xmax=10,
            ymax=4.0]
            \addplot+[mark=none] table [x=step, y=inception_score, col sep=comma] {data/inceptionScores.csv};
            \end{axis}
		\end{tikzpicture}%
        
        \hspace{1cm}(a) \hspace{5.5cm} (b)
	
	\caption{ (a) Classification accuracy for chunks and (b) inception score for chunks}
	\label{fig:chunkResults}
\end{figure*}
\subsubsection{Effect of Ranking}
To check the suitability of $D'$ to rank the generated samples, we further analyze the behavior of the ranked samples. Therefore, we split 30 ranked samples into chunks of size 3, yielding 10 ranked chunks. In a first experiment, we evaluate the class-discriminativeness of those chunks. In this regard, we use the initialized $D'$ and consider every single chunk as test set. The classification results in top-1-accuracy and top-5-accuracy are shown in Fig.~\ref{fig:chunkResults}a. It can be observed that the classification accuracy for high-ranked chunks is higher in test time. This shows the ranking is working in the desired fashion, i.e. prioritizing generated samples for which $D'$ is more confident w.r.t. to correct class-discriminative information. 
In a second experiment, we study the quality of the generated images across the ranked chunks. Note that ranking is performed based on class-discriminativeness alone and specifically does not take into account the quality of the image. Therefore, we use the inception score \cite{salimans_improved_2016} 
to assess the quality of images generated by GANs. 
As can be observed from Fig.\ref{fig:chunkResults}b, the quality of the images remains constant across the chunks, validating that we do not suffer a quality drop by ranking images based on class-discriminativeness. Hence, our proposed ranking allows to pick images that perform best for classification without any loss in quality of the generated images. Fig. \ref{fig:lastFigure} shows some generated images for one category ranked in different iterations. We can observe that the ranking improves over the iterations.

\section{Conclusion and Future Work}
In this paper, we proposed to extend few-shot learning to deal with multimodal data and introduced a discriminative tcGAN for sample generation along with a self-paced strategy for sample selection. Experiments on CUB and Oxford-102 demonstrate that learning generative models in a cross-modal fashion facilitates few-shot learning by compensating the lack of data in novel categories. For future work we plan to incorporate class-discriminativeness property at the representation learning stage jointly with the ranking loss of the self-paced stage. Furthermore, we seek to investigate the use of self-paced GAN for other cross-modal generative tasks. Moreover, future research will be conducted on optimizing the text embedding in context of multimodality. Last, a more generalized task setup in which classification is performed on $\Cbase\cup\Cnovel$, similar to the task proposed in \cite{XLSA18} for zero-shot learning, is a valuable research direction.

{\small
\bibliographystyle{ieee}
\bibliography{main}

\begin{thebibliography}{10}\itemsep=-1pt

\bibitem{bengio2009curriculum}
Y.~Bengio, J.~Louradour, R.~Collobert, and J.~Weston.
\newblock Curriculum learning.
\newblock In {\em ICML}, pages 41--48, 2009.

\bibitem{bertinetto_learning_2016}
L.~Bertinetto, J.~F. Henriques, J.~Valmadre, P.~Torr, and A.~Vedaldi.
\newblock Learning feed-forward one-shot learners.
\newblock In {\em Advances in {Neural} {Information} {Processing} {Systems}},
  pages 523--531, 2016.

\bibitem{bromley1994signature}
J.~Bromley, I.~Guyon, Y.~LeCun, E.~S{\"a}ckinger, and R.~Shah.
\newblock Signature verification using a" siamese" time delay neural network.
\newblock In {\em Advances in Neural Information Processing Systems}, pages
  737--744, 1994.

\bibitem{chen2015webly}
X.~Chen and A.~Gupta.
\newblock Webly supervised learning of convolutional networks.
\newblock In {\em ICCV}, pages 1431--1439, 2015.

\bibitem{douze_low-shot_2017}
M.~Douze, A.~Szlam, B.~Hariharan, and H.~J{\'e}gou.
\newblock Low-shot learning with large-scale diffusion.
\newblock {\em CoRR}, 2017.

\bibitem{faghri_vse++:_2017}
F.~Faghri, D.~J. Fleet, J.~R. Kiros, and S.~Fidler.
\newblock {VSE}++: {Improving} {Visual}-{Semantic} {Embeddings} with {Hard}
  {Negatives}.
\newblock {\em arXiv:1707.05612 [cs]}, July 2017.
\newblock arXiv: 1707.05612.

\bibitem{goodfellow_generative_2014}
I.~Goodfellow, J.~Pouget-Abadie, M.~Mirza, B.~Xu, D.~Warde-Farley, S.~Ozair,
  A.~Courville, and Y.~Bengio.
\newblock Generative adversarial nets.
\newblock In {\em NIPS}, pages 2672--2680, 2014.

\bibitem{hariharan_low-shot_2017}
B.~Hariharan and R.~Girshick.
\newblock Low-shot {Visual} {Recognition} by {Shrinking} and {Hallucinating}
  {Features}.
\newblock In {\em {ICCV}}, 2017.

\bibitem{karpathy_deep_2015}
A.~Karpathy and L.~Fei-Fei.
\newblock Deep visual-semantic alignments for generating image descriptions.
\newblock In {\em CVPR}, pages 3128--3137, 2015.

\bibitem{kingma2014adam}
D.~P. Kingma and J.~Ba.
\newblock Adam: A method for stochastic optimization.
\newblock {\em arXiv preprint arXiv:1412.6980}, 2014.

\bibitem{kingma2013auto}
D.~P. Kingma and M.~Welling.
\newblock Auto-encoding variational bayes.
\newblock {\em arXiv preprint arXiv:1312.6114}, 2013.

\bibitem{kiros_unifying_2014}
R.~Kiros, R.~Salakhutdinov, and R.~S. Zemel.
\newblock Unifying {Visual}-{Semantic} {Embeddings} with {Multimodal} {Neural}
  {Language} {Models}.
\newblock {\em arXiv:1411.2539 [cs]}, Nov. 2014.
\newblock arXiv: 1411.2539.

\bibitem{koch_siamese_2015}
G.~Koch, R.~Zemel, and R.~Salakhutdinov.
\newblock Siamese neural networks for one-shot image recognition.
\newblock In {\em ICML Deep Learning Workshop}, volume~2, 2015.

\bibitem{kumar2010self}
M.~P. Kumar, B.~Packer, and D.~Koller.
\newblock Self-paced learning for latent variable models.
\newblock In {\em NIPS}, pages 1189--1197, 2010.

\bibitem{KwakZ16a}
H.~Kwak and B.~Zhang.
\newblock Ways of conditioning generative adversarial networks.
\newblock {\em CoRR}, abs/1611.01455, 2016.

\bibitem{liang2015towards}
X.~Liang, S.~Liu, Y.~Wei, L.~Liu, L.~Lin, and S.~Yan.
\newblock Towards computational baby learning: A weakly-supervised approach for
  object detection.
\newblock In {\em ICCV}, pages 999--1007, 2015.

\bibitem{DBLP:journals/corr/MirzaO14}
M.~Mirza and S.~Osindero.
\newblock Conditional generative adversarial nets.
\newblock {\em CoRR}, abs/1411.1784, 2014.

\bibitem{mishra2017generative}
A.~Mishra, M.~Reddy, A.~Mittal, and H.~A. Murthy.
\newblock A generative model for zero shot learning using conditional
  variational autoencoders.
\newblock {\em arXiv preprint arXiv:1709.00663}, 2017.

\bibitem{nilsback2008automated}
M.-E. Nilsback and A.~Zisserman.
\newblock Automated flower classification over a large number of classes.
\newblock In {\em ICVGIP}, pages 722--729. IEEE, 2008.

\bibitem{odena2016conditional}
A.~Odena, C.~Olah, and J.~Shlens.
\newblock Conditional image synthesis with auxiliary classifier gans.
\newblock {\em arXiv preprint arXiv:1610.09585}, 2016.

\bibitem{ravi_optimization_2017}
S.~Ravi and H.~Larochelle.
\newblock Optimization as a model for few-shot learning.
\newblock In {\em {InternationalConference} on {Learning} {Representations}},
  2017.

\bibitem{reed_learning_2016}
S.~Reed, Z.~Akata, H.~Lee, and B.~Schiele.
\newblock Learning deep representations of fine-grained visual descriptions.
\newblock In {\em CVPR}, pages 49--58, 2016.

\bibitem{reed16_gen}
S.~Reed, Z.~Akata, X.~Yan, L.~Logeswaran, B.~Schiele, and H.~Lee.
\newblock Generative adversarial text to image synthesis.
\newblock In {\em ICML}, volume~48 of {\em Proceedings of Machine Learning
  Research}, pages 1060--1069. PMLR, 2016.

\bibitem{salimans_improved_2016}
T.~Salimans, I.~Goodfellow, W.~Zaremba, V.~Cheung, A.~Radford, and X.~Chen.
\newblock Improved techniques for training gans.
\newblock pages 2234--2242, 2016.

\bibitem{sangineto2016self}
E.~Sangineto, M.~Nabi, D.~Culibrk, and N.~Sebe.
\newblock Self paced deep learning for weakly supervised object detection.
\newblock {\em IEEE Transactions on Pattern Analysis and Machine Intelligence},
  2018.

\bibitem{sharma2018chatpainter}
S.~Sharma, D.~Suhubdy, V.~Michalski, S.~E. Kahou, and Y.~Bengio.
\newblock Chatpainter: Improving text to image generation using dialogue.
\newblock {\em arXiv preprint arXiv:1802.08216}, 2018.

\bibitem{snell_prototypical_2017}
J.~Snell, K.~Swersky, and R.~Zemel.
\newblock Prototypical networks for few-shot learning.
\newblock In {\em NIPS}, pages 4080--4090. 2017.

\bibitem{supancic2013self}
J.~S. Supancic~III and D.~Ramanan.
\newblock Self-paced learning for long-term tracking.
\newblock In {\em CVPR}, pages 2379--2386, 2013.

\bibitem{taigman2014deepface}
Y.~Taigman, M.~Yang, M.~Ranzato, and L.~Wolf.
\newblock Deepface: Closing the gap to human-level performance in face
  verification.
\newblock In {\em CVPR}, pages 1701--1708, 2014.

\bibitem{vinyals_matching_2016}
O.~Vinyals, C.~Blundell, T.~Lillicrap, D.~Wierstra, et~al.
\newblock Matching networks for one shot learning.
\newblock In {\em NIPS}, pages 3630--3638, 2016.

\bibitem{WahCUB_200_2011}
C.~Wah, S.~Branson, P.~Welinder, P.~Perona, and S.~Belongie.
\newblock {The Caltech-UCSD Birds-200-2011 Dataset}.
\newblock Technical report, 2011.

\bibitem{wang_low-shot_2018}
Y.-X. Wang, R.~Girshick, M.~Hebert, and B.~Hariharan.
\newblock {Low-Shot Learning from Imaginary Data}.
\newblock In {\em {CVPR}}, 2018.

\bibitem{XLSA18}
Y.~Xian, H.~C. Lampert, B.~Schiele, and Z.~Akata.
\newblock Zero-shot learning - a comprehensive evaluation of the good, the bad
  and the ugly.
\newblock {\em TPAMI}, 2018.

\bibitem{xu_attngan:_2017}
T.~Xu, P.~Zhang, Q.~Huang, H.~Zhang, Z.~Gan, X.~Huang, and X.~He.
\newblock {AttnGAN}: {Fine}-{Grained} {Text} to {Image} {Generation} with
  {Attentional} {Generative} {Adversarial} {Networks}.
\newblock {\em arXiv:1711.10485 [cs]}, Nov. 2017.
\newblock arXiv: 1711.10485.

\bibitem{yan2016attribute2image}
X.~Yan, J.~Yang, K.~Sohn, and H.~Lee.
\newblock Attribute2image: Conditional image generation from visual attributes.
\newblock In {\em ECCV}, 2016.

\bibitem{yoo_efficient_2017}
D.~{Y}oo, H.~Fan, V.~N. Boddeti, and K.~M. Kitani.
\newblock Efficient {K}-{Shot} {Learning} with {Regularized} {Deep} {Networks}.
\newblock In {\em AAAI}, 2018.

\bibitem{zhang2017bridging}
D.~Zhang, D.~Meng, L.~Zhao, and J.~Han.
\newblock Bridging saliency detection to weakly supervised object detection
  based on self-paced curriculum learning.
\newblock {\em arXiv preprint arXiv:1703.01290}, 2017.

\bibitem{zhang_stackgan++:_2017}
H.~Zhang, T.~Xu, H.~Li, S.~Zhang, X.~Wang, X.~Huang, and D.~Metaxas.
\newblock Stackgan: Text to photo-realistic image synthesis with stacked
  generative adversarial networks.
\newblock In {\em {ICCV}}, 2017.

\end{thebibliography}
}

\end{document}